\documentclass[10pt,twocolumn,letterpaper]{article}

\usepackage{cvpr}
\usepackage{times}
\usepackage{epsfig}
\usepackage{graphicx}
\usepackage{amsmath}
\usepackage{amssymb}

\usepackage{algorithm}
\usepackage{listings}

\usepackage[pagebackref=true,breaklinks=true,letterpaper=true,colorlinks,bookmarks=false]{hyperref}

\cvprfinalcopy 

\def\cvprPaperID{11705} 

\ifcvprfinal\fi
\begin{document}

\title{Dynamic Class Queue for Large Scale Face Recognition In the Wild}

\author{Bi Li$^{1, 2}$\thanks{Corresponding to Bi Li and Teng Xi. Equal contribution.},  
	Teng Xi$^{1, 3*}$, 
	Gang Zhang$^{1}$, 
	Haocheng Feng$^{1}$, 
	Junyu Han$^{1}$,
	Jingtuo Liu$^{1}$,
    Errui Ding$^{1}$,
    Wenyu Liu$^{2}$ \\
    $^{1}$Department of Computer Vision
Technology (VIS), Baidu Inc.\\
	$^{2}$School of Electronic Information and Communications, Huazhong University of Science and Technology\\
	$^{3}$Department of Computer Science and Technology, Tsinghua University\\
	\tt\small{$^{1}$\{libi01,xiteng01,zhanggang03,fenghaocheng,hanjunyu,liujingtuo,dingerrui\}@baidu.com}\\
	\tt\small{$^{2}$liuwy@hust.edu.cn}}

\maketitle
\thispagestyle{empty}

\begin{abstract}
   Learning discriminative representation using large-scale face datasets in the wild is crucial for real-world applications, yet it remains challenging. The difficulties lie in many aspects and this work focus on computing resource constraint and long-tailed class distribution. Recently, classification-based representation learning with deep neural networks and well-designed losses have demonstrated good recognition performance. However, the computing and memory cost linearly scales up to the number of identities (classes) in the training set, and the learning process suffers from unbalanced classes. In this work, we propose a dynamic class queue (DCQ) to tackle these two problems. Specifically, for each iteration during training, a subset of classes for recognition are \textbf{dynamically selected} and their class weights are \textbf{dynamically generated} on-the-fly which are stored in a queue. Since only a subset of classes is selected for each iteration, the computing requirement is reduced. By using a single server without model parallel, we empirically verify in large-scale datasets that 10\% of classes are sufficient to achieve similar performance as using all classes. Moreover, the class weights are dynamically generated in a few-shot manner and therefore suitable for tail classes with only a few instances. We show clear improvement over a strong baseline in the largest public dataset Megaface Challenge2 (MF2) which has 672K identities and over 88\% of them have less than 10 instances. Code is available at \url{https://github.com/bilylee/DCQ}

\end{abstract}

\begin{figure}[t]
   \centering
   \includegraphics[width=0.8\linewidth]{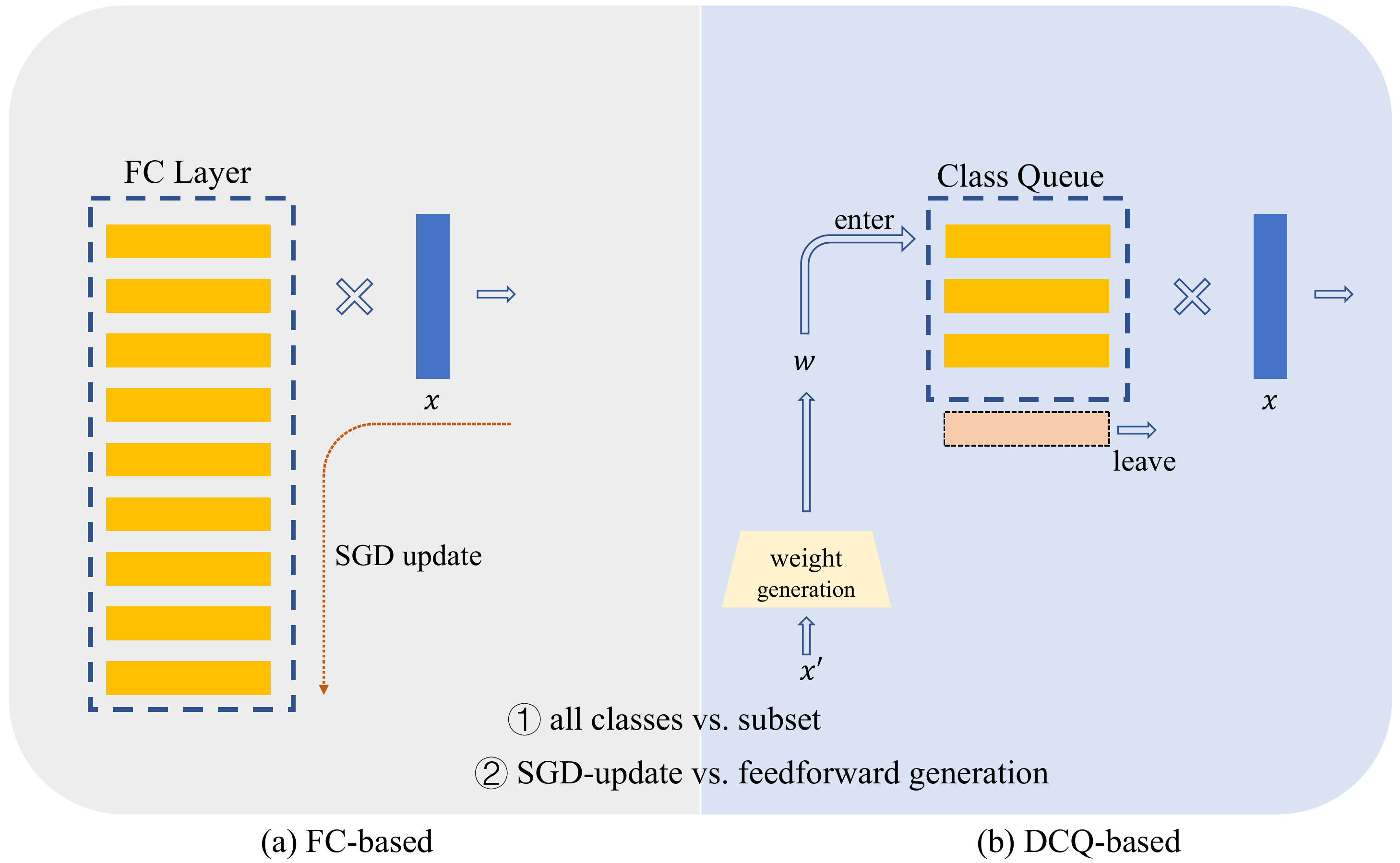}
   \caption{High-level comparison between the fully connected layer (FC layer) used in classification and the proposed DCQ module. Generally, there are two differences. 1) For the FC layer, all classes in the training dataset are included in the FC layer (each row in the FC layer represents a class weight). While for DCQ, only a subset is used. 2) The class weights in the FC layer are randomly initialized and then updated via SGD. In contrast, DCQ gets the class weights in a few-shot manner based on another instance $x'$ with the same identity as input $x$.\label{fig:intro}}
\end{figure}

\section{Introduction}
Recently, face recognition has witnessed great progress along with the development of deep neural networks and large-scale datasets. The size of the public training set has been steadily increasing from CASIA-Webface\cite{yi2014learning} (10K identity, 0.5M instance) $\rightarrow$ MS-Celeb-1M\cite{guo2016ms} (100K identity, 5M instance) $\rightarrow$ MF2\cite{nech2017level} (672K identity, 4.7M instance). For commercial applications, the training dataset easily scales up to millions of identities and it is a matter of time to reach billions of identities.

More data brings better performance. However, the training difficulty accumulates along with the growth of the training data. First of all, it simply needs more computing resources. For classification-based methods, where each identity is taken as a class and the feature extractor is learned through the classification task, the memory consumption of the fully connected layer $W\in\mathbb{R}^{D\times C}$ linearly scales up to the number of identities $C$ in the training set. So is the cost to compute the matrix multiplication between the FC layer and the input feature. Secondly, for data gathered in the real world, the class distribution is typically long-tailed, that is, some classes have abundant instances (head classes) while most classes have few instances (tail classes). For example, the MF2 dataset contains images gathered from Flickr, and over 88\% of identities have less than 10 images. As witnessed by various studies\cite{branco2016survey,zhang2017range,zhong_unequal-training_2019}, long-tailed classification is itself a challenging problem. 

To tackle the computing resource constraint, one option is to adopt pairwise-based methods which have the benefits of being class-number agnostic and therefore can be potentially extended to datasets with an arbitrary number of identities. However, the pair sampling mechanism is critical for this method to achieve good performance\cite{schroff2015facenet} and it takes a much longer time to converge.
Another option is to dynamically reduce the number of classes used during training. Zhang et al.\cite{zhang_accelerated_2018} propose to use a hashing forest to partition the space of class weights into small cells.
Given a training sample $x$, they walk through the forest to find the closest cell and use classes within as the FC Layer. 
Concurrent with our work, An et al.\cite{an2020partial} demonstrate that 
randomly sampling classes from all classes can also achieve matching performance as using all classes. These methods can largely reduce the computational cost by using a subset of classes for computing the matrix multiplication and the softmax function. However, they still require all class weights to be stored in the memory. \cite{zhang_accelerated_2018} uses a parameter server to overcome this problem while \cite{an2020partial} uses model parallel\footnote{It splits the model into parts and places them in different GPUs to meet memory constraints.} to distribute the class weights into several GPUs. Moreover, since the class weights $W$ are updated by SGD, they also need to store all optimization-related stats of the class weights such as the momentum in memory. 

As for long-tailed class distribution, a simple solution would be removing tail classes such that the class distribution is balanced. However, the full potential of the training data is undermined in this way. Another option is to use class-based sampling during training, i.e. sampling classes with equal probability\cite{chawla2002smote}. However, as demonstrated in our experiment, this is not necessary and even harmful for our method. Zhong et al.\cite{zhong_unequal-training_2019} propose a two-stage training mechanism where the model is firstly trained with only the head classes and then finetuned with both head and tail classes. As noted by the authors, the second stage needs careful manual tuning to achieve good performance. It is of interest to design methods that require only single-stage training. 

In this work, we propose a single-stage method that is computationally efficient and has low memory consumption. The key innovation is to design a dynamic class queue. The meaning of ``dynamic'' is two-fold. First, the class subset used for classification is dynamically selected. The computational cost is reduced since only a part of the classes are involved in the computation. Second, the class weights are dynamically generated on-the-fly instead of learning via SGD. Since the class weight is dynamically generated at each iteration, it does not require storing all class weights or optimization-related stats in the memory. These class weights are stored in a queue where the queue size can be 10x smaller than all classes in our experiments. Importantly, the class weights are dynamically generated in a few-shot way which is friendly to tail classes and therefore helpful for training long-tailed datasets. For a visual illustration of the proposed method, please refer to Figure~\ref{fig:intro}.

We empirically verify that the proposed method is effective and efficient. By using a single server and less than 9GB memory per GPU card without model parallel, the proposed method uses 10\% of classes while still achieving similar performance as the baseline which uses all classes in the large-scale dataset (MS1MV2\cite{deng2019arcface}). The proposed method is most useful in the real-world long-tailed dataset. We demonstrate this on the MF2\cite{nech2017level} dataset which outperforms a strong baseline by 0.75\% (absolute change) in identification and 1.72\% in verification tasks with only 10\% classes. 

\begin{figure*}[t]
   \centering
   \includegraphics[width=0.7\linewidth]{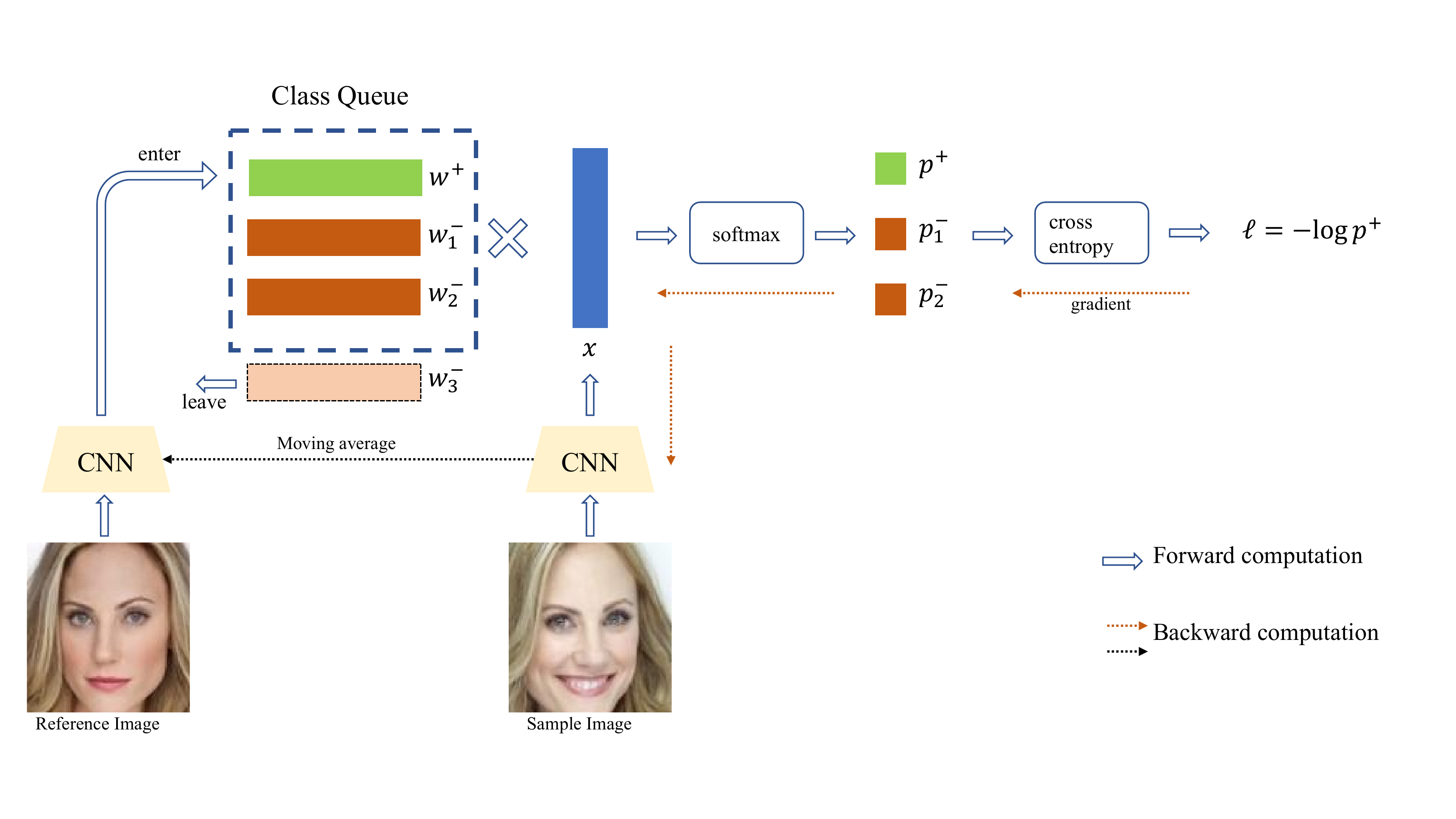}
   \caption{\textbf{The framework of the proposed method.} Given two images of the same identities, the reference image passes through the weight generator to get the class weight and update the class queue. The sample image is embedded by the feature extractor, then classified by the class queue and supervised by the classification loss. During backward pass, the gradients update the feature extractor. The weight generator is updated by the moving average of the feature extractor.\label{fig:framework}}
\end{figure*}

\section{Related Work}
\textbf{Pairwise-loss based.} A straightforward way to learn the representation is by sampling pair or triplet samples\cite{chopra2005learning,hoffer2015deep,wang2014learning} and minimize the pairwise losses. The simplest form of pairwise loss is verification loss which samples two images at a time and minimizes the $L_2$ distance when the two images come from the same identity, otherwise enforces a margin. 
FaceNet\cite{schroff2015facenet} adopts triplet loss which samples three images at a time, an anchor image $A$, a positive image $P$ and a negative image $N$, and the relative distance $d(A, P) - d(A,N)$ should be minimized below a margin. 
Our work can be interpreted in the pairwise-loss-based framework. The difference is, 
instead of using only negatives from the current batch, we build a class queue that is much larger than the batch and is dynamically updated by the samples in the batch.

\textbf{Classification based.} Recently, classification based representation learning have been actively exploited in face recognition\cite{wen2016discriminative,liu2017sphereface,wang2018additive,wang2018cosface,deng2019arcface}.
To ensure intra-class compactness, \cite{wen2016discriminative} proposes the center loss to penalize the distance between the image feature and the corresponding class center. For inter-class dispersion, the feature extractor is also constrained by the entropy loss. SphereFace\cite{liu2017sphereface,liu2016large} notice that the entropy loss can be reinterpreted in a geometric way by $l_2$ normalizing the weights of the last FC layer and therefore projecting the distance metric to angular space. An angular margin is then introduced to enforce inter-class dispersion. CosFace\cite{wang2018cosface} further normalizes both the weight and the features to remove radial variations and introduces a cosine margin. Compared to SphereFace\cite{liu2017sphereface}, CosFace is more robust during training since it overcomes the optimization difficulty in the angular space. ArcFace\cite{deng2019arcface} also notices the training difficulty of SphereFace and proposes an additive angular margin.

\textbf{Momentum-based model update.} Feature drift has been one of the key challenges to maintain a memory(queue) that works across the batch. Recently, He et al.\cite{he_momentum_2020} propose a momentum-based method to keep the features consistent in the memory and largely reduced the gap between supervised and unsupervised learning. Du et al.\cite{du_semi-siamese_2020} notice the shallow face problem in face recognition and propose to use the momentum-based update to keep the class weights different from sample features. Their method bears similarities with our work in that class weights are generated in a feedforward manner based on a support image. Nevertheless, the main goal of this work is to design methods for training with large-scale datasets in the wild, therefore we focus on the long-tailed problem and propose to use a subset of classes for large-scale training.

\newcommand{\algcomment}[1]{%
    \noindent%
    {\footnotesize #1\par}%
}

\begin{algorithm}[t]
\caption{Pseudocode of DCQ.}
\label{alg:code}
\algcomment{\fontsize{7.2pt}{0em}\selectfont \texttt{bmm}: batch matrix multiplication; \texttt{mm}: matrix multiplication; \texttt{cat}: concatenate.
}
\definecolor{codeblue}{rgb}{0.25,0.5,0.5}
\lstset{
  backgroundcolor=\color{white},
  basicstyle=\fontsize{7.2pt}{7.2pt}\ttfamily\selectfont,
  columns=fullflexible,
  breaklines=true,
  captionpos=b,
  commentstyle=\fontsize{7.2pt}{7.2pt}\color{codeblue},
  keywordstyle=\fontsize{7.2pt}{7.2pt},
}
\begin{lstlisting}[language=python]
# f & g: feature extractor & weight generator network
# w_queue: class weights queue of length K (CxK)
# l_queue: class label queue of length K (1xK)
# a: momentum
# s & m: scale and margin in CosFace loss
g.params = f.params  # initialize
for (x_t, x_w, y) in loader:
    # x_t: test sample, x_w: reference sample
    # y: class labels of x_t
    t = f.forward(x_t)  # test features: NxC
    w = g.forward(x_w)  # class weights: NxC
    w = w.detach()  # no gradient to class weights

    # positive and negative logits
    l_pos = bmm(t.view(N,1,C), w.view(N,C,1)) # NX1
    l_neg = mm(q.view(N,C), w_queue.view(C,K)) # NxK
    
    # mute duplicate responses in the queue
    l_diff = y - l_queue  # N x K
    l_neg = l_neg.masked_fill(l_diff == 0, -1e9)
    
    # CosFace Loss
    l_pos = l_pos - m  # add margin
    logits = cat([l_pos, l_neg], dim=1)
    logits *= s # add scale 
    labels = zeros(N)  # positives are the 0-th
    loss = CrossEntropyLoss(logits, labels)

    # SGD update and momentum update
    loss.backward()
    update(f.params)
    g.params = a*g.params+(1-a)*f.params

    # update queue, first in first out
    enqueue(w_queue, w)
    enqueue(l_queue, y)
    dequeue(w_queue)
    dequeue(l_queue)
\end{lstlisting}
\end{algorithm}

\section{Method}
In this section, we first introduce the preliminaries (Sec.~\ref{sec:preliminaries}) and introduce the difficulties faced in training large-scale datasets (Sec.~\ref{sec:difficulty}), then introduce the proposed dynamic class queue in detail (Sec.~\ref{sec:dcq}). For a visual illustration of the proposed method, please refer to Figure.~\ref{fig:framework}.

\subsection{Preliminaries}\label{sec:preliminaries}
\textbf{Classification-based Representation Learning.} 
Given a training dataset $T \triangleq \{(x_i, y_i), i \in \{1,2,\cdots,N\}\}$, where $N=|T|$ is number of images in $T$, $x_i \in \mathbb{R}^{H\times W \times 3}$ is a face image and $y_i \in \mathcal{Y} = \{1, 2, .., C\}$ is the corresponding face identity encoded via one-hot vector. $C$ is the total number of identities which can be over millions in large-scale datasets. To learn a feature extractor $\varphi_\phi(\cdot)$ with learnable parameters $\phi$, images are first encoded by the feature extractor $f = \varphi_\phi(x) \in \mathbb{R}^D$. Then, they are classified by a linear layer (fully connected layer) with weight $W \in \mathbb{R}^{D\times C}$, that is $\hat{y} = W^T \varphi_\phi(x)$. The feature extractor is trained by minimizing the loss: $\arg\min_{\phi, W} \frac{1}{N} \sum_{i=1}^N \mathcal{L}(W^T \varphi_\phi(x_i), y_i)$ where $\mathcal{L}(\hat{y}, y)$ measures the discrepancy between the predicted value $\hat{y}$ and the groundtruth $y$. Cross entropy loss is typically used for classification task. However, it is found that a variant of cross entropy loss which first projecting the feature $f$ and the linear classifier $W$ into the spherical space and add large cosine margin, the model can learn more discriminative features for face recognition. Specifically, in this work, we adopt the loss function used in CosFace\cite{wang2018cosface}, that is:

\begin{equation} \label{eq:loss}
   \mathcal{L} = - \ln \frac{e^{s(\cos(\theta_y) -m)}}{e^{s(\cos(\theta_y) - m) } + \sum_{j \in \mathcal{Y} / \{y\}}e^{s\cos(\theta_j)}}
\end{equation}

where $\cos(\theta) = (\frac{W}{||W||})^T \frac{f}{||f||} \in \mathbb{R}^C$, $\cos(\theta_j)$ is the $j$-th value. $\mathcal{Y} / \{y\}$ are all classes excluding the groundtruth $y$. $s$ and $m$ are hyperparameters, $s$ controls the scale and $m$ is the cosine margin. Note that the proposed method is not limited to CosFace and can be applied to any other loss functions such as ArcFace\cite{deng2019arcface}.

The loss is minimized via stochastic gradient descent (SGD). During evaluation, the FC layer is removed and only the feature extractor is used. The learned representation can be used for either face verification by thresholding the distance between the test and the reference face, or face identification by searching the nearest neighbor in a face database.

\textbf{Feedforward weight generation} The class weights $w$ in the FC layer are randomly initialized and iteratively updated via SGD. This is beneficial for classes with sufficient training samples. However, it struggles with tail classes. In this work, we would like to generate the class weights in a feedforward manner and on-the-fly at each iteration. Moreover, it should be friendly to tail classes with few instances. We tackle this problem by following the few-shot learning or meta-learning methodology. Given a support image (or reference image) $x'$ from the same identity as the query image $x$, it learns a weight generator function $g(\cdot)$ such that $w = g(x')$.

\subsection{Training Difficulties for large-scale Datasets} \label{sec:difficulty}
\textbf{Hardware constraints.} In the above classification-based framework, the size of the classifier weight $W\in \mathbb{R}^{D\times C}$ linearly increases with the number of identities in the training dataset $C$. This can be problematic for large-scale face recognition with a large number of identities. The FC layer easily exceeds the memory limit of the GPU and solely computes the $\hat{y} = W^T \varphi_\phi(x)$ will dominate the computation cost.

\textbf{Learning Difficulties.} Even if the computing and memory resource is unlimited, the learning process faces intrinsic difficulties. To illustrate, we use cross entropy loss as an example. The reasoning holds for other cross entropy based variants such as CosFace and ArcFace, et al. 

During training, given a sample $x$, we can get its predictions $\hat{y} = W^T \varphi_\phi(x) \in \mathbb{R}^{C}$.  The predictions are then normalized into probabilities by a softmax function: $p_i = \frac{e^{\hat{y}_i}}{\sum_{j}e^{\hat{y}_j}}$ which has properties $\sum_{i=1}^{C} p_i = 1$ and $p_i \ge 0$.
Denote the probability of the groundtruth as $p^+$ and others as $p^-_j$. Without loss of generality, we have $p = [p^+, p^-_1, p^-_2, ..., p^-_{C-1}]$. Correspondingly, the FC layer can be represented as $W = [w^+, w^-_1, w^-_2, ..., w^-_{C-1}]$. The cross entropy loss of the predictions is then $ \mathcal{L}_{CE} = -\log p^+$. Notice that even though only $p^+$ is used in the cross entropy loss, the predictions of other classes are implicitly contained in the denominator of $p^+$ due to the softmax function.

Then, the gradients of the cross entropy loss $\mathcal{L}$ with respect to the feature $f$ and the weights in the FC layer are:

\begin{align}\label{eq:gradient}
   \frac{\partial \mathcal{L}}{\partial f} &= -(1-p^+)w^+ + \sum_{i}{p^-_i w^-_i} \\
   \frac{\partial \mathcal{L}}{\partial w^+} &= -(1-p^+)f \\
   \frac{\partial \mathcal{L}}{\partial w^-_j} &= p^-_j f
\end{align}   
Intuitively, these gradients show the ``pull'' and ``push'' forces caused by the positive and negative class weights and samples. For example, the feature $f$ is updated by the total forces of positive and negative classes. The positive class weights $w^+$ are pulled to the feature $f$ while the negative class weights $w^-$ are pushed away from $f$. The probabilities $p$ indicates the strength of these forces. Interestingly, the gradient of the feature is balanced since $(1-p^+) = \sum_{i}{p^-_i}$. 

Now lets focus on the update of a single class weight $w$. Through the training process, it is pushed away by samples from other classes and pulled close to its instances, that is,

\begin{equation}
   w = w_0 + \sum_{a\in C^+}(1-p^+_a)f_a - \sum_{b\in C^-}p^-_bf_b
\end{equation}
Where $C^+$ represents all instances belong to this class accoutered during training and $C^-$ represents instances of other classes. $w_0$ is the random initialization. Learning rates are omitted. 

This can be problematic for large-scale datasets in the wild. For large-scale datasets, the classes are typically long-tailed. That is, some classes have abundant instances while most classes contain only few instances. Therefore, for tail classes with only few instances, their class weights are dominated by push-away updates and cannot represents their instances.

\begin{figure}[t]
   \centering
   \includegraphics[width=0.4\linewidth]{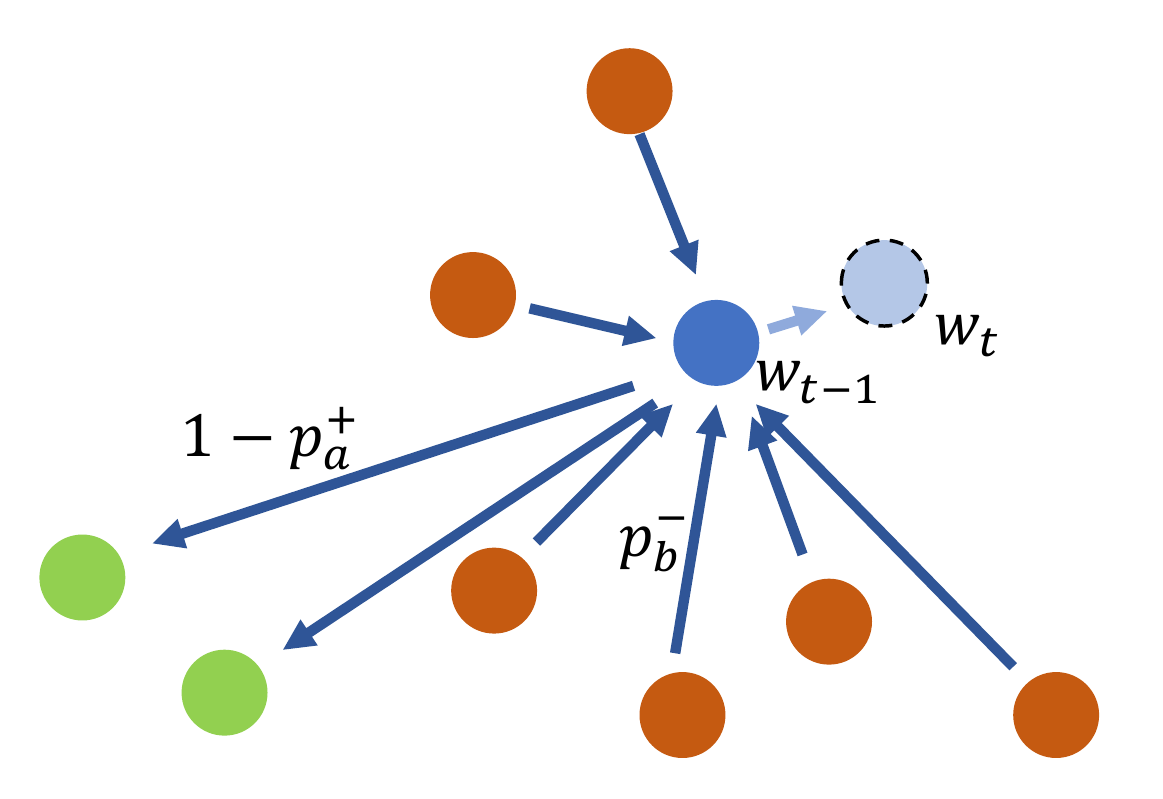}
   \caption{\textbf{The illustration of the learning difficulty.} \label{fig:weight_update}}
\end{figure}

\subsection{Dynamic Class Queue} \label{sec:dcq}
In this work, we design a dynamic class queue module to tackle all the above difficulties: 1) computational and 2) memory cost of using all classes and 3) long-tailed class distribution. Two crucial points of the proposed method are: First, the classes used for classification are dynamically selected. Second, the class weights are dynamically generated on-the-fly instead of randomly initialized and then updated via SGD. Since only a subset of classes is used at each iteration, the computational cost is largely reduced. Moreover, the class weights are dynamically generated at each iteration and therefore do not require storing all class weights in the memory. Finally, the class weights are generated in a few-shot manner and friendly to tail classes.

\textbf{Dynamic Class Selection.} The key idea behind dynamic class selection is that since the model is optimized via stochastic gradient descent and the training process proceeds in a batch-wise way, therefore not all classes are needed in one batch. Let $\mathcal{X}_B = \{x_1, x_2, ..., x_B\}$ be a batch of images during training, and $\mathcal{Y}_B$ be their corresponding identity labels. Let $\mathcal{Y}_S$ be a set of classes used for classification, it is clear that $\mathcal{Y}_B \subseteq \mathcal{Y}_S \subseteq \mathcal{Y}$. Accordingly, we can rewrite eq.~\ref{eq:loss} by substituting all classes $\mathcal{Y}$ with a subset $\mathcal{Y}_S$:

\begin{equation} \label{eq:subset-loss}
   \mathcal{L}_S = - \ln \frac{e^{s(\cos(\theta_y) -m)}}{e^{s(\cos(\theta_y) - m) } + \sum_{j \in \mathcal{Y}_S / \{y\}}e^{s\cos(\theta_j)}}
\end{equation}

Now, the question is how to design the class set $\mathcal{Y}_S$? A trivial solution will be setting $\mathcal{Y}_S = \mathcal{Y}$, that is, using all classes in the training set. However, to overcome the scaling problem, we would like a class set that is much smaller than all classes, i.e., $|\mathcal{Y}_S| \ll |\mathcal{Y}|$. Since the classes of images in the batch $\mathcal{Y}_B$ change for every batch, it means $\mathcal{Y}_S$ should be dynamically updated every iteration during training. We call $\mathcal{Y}_S$ the dynamic class set.

 Concurrent with our work, An et al.\cite{an2020partial} indicates that random sampling a subset of classes performs as well as using all classes. Let $\mathcal{Y}_S^t$ be the dynamic class set at iteration $t$, by using random sampling we have $\mathcal{Y}_S^t = \mathcal{Y}_B^t \cup \text{Sample}(\mathcal{Y}/\mathcal{Y}_B^t, K).$ Further more,we notice that instead of using totally different classes at each iteration, we can build a queue with size $B + K$ and update $B$ element of the queue at each iteration. That is,
\begin{equation}
   \mathcal{Y}_S^t = \mathcal{Y}_B^{t} \cup \mathcal{Y}_B^{t-1} \cup \mathcal{Y}_S^{t-1} / \mathcal{Y}_B^{\text{oldest}}.
\end{equation}
where $\mathcal{Y}_B^{\text{oldest}}$ is the oldest batch in the queue (i.e., first in first out). Since $\mathcal{Y}_B^{t}$ is randomly sampled at each batch, therefore the classes in the queue can be seen as a random sampling from all classes. 

The queue-based dynamic class selection is important in our work because the class weights are dynamically generated. By reusing the $K$ class weights in the queue (only $B$ class weights are updated), the computation for getting all the selected class weights is greatly reduced. However, two problems need to be solved to make it work. One is that duplicate classes may exist in the queue. We overcome this problem by setting the logits to $-\infty$ for duplicate classes (Refer to Algorithm 1 for details). Another problem with reusing the class weights is that these weights are generated at earlier iterations and may have feature drift. We will tackle this problem in the following subsection.

\textbf{Dynamic Class Generation.} There are various methods in the meta-learning literature to design the mapping function, we will adopt the simplest one in this work which already demonstrates good performance empirically. Note that the design of the mapping function is orthogonal to other components in this work and we expect better performance with the development in the meta-learning field. Specifically, we adopt the siamese network and therefore $g(\cdot) = \varphi_\phi(\cdot)$ where $\varphi_\phi(\cdot)$ is the feature extractor. We have also tried prototypical network which takes the mean of several support samples as the class weight, i.e. $w = \frac{1}{n}\sum_{i=1}^{n}\varphi_\phi(x'_i)$, however it does not demonstrate better performance. We emphasize that class weight generated via meta learning is friendly to tail classes as it is agnostic to the number of samples in the class.

One problem with the current class weight generation method is that since we reuse the $K$ class weights in the queue which are generated in earlier iterations during training, they may have different feature distribution compared to the recently generated class weights. Therefore, the model can easily discriminate between the positive classes (which are generated in the current batch) and the negative classes (which are reused in the queue) and the learned feature is not discriminative. To overcome this problem, we need to bridge the feature gap between class weights generated at different time steps. Recently, He et al.\cite{he_momentum_2020} proposes to mitigate the feature drift problem by taking the moving average of the neural network parameters. We adopt this method in this work. Specifically, at one iteration, the feature extractor is first updated via SGD, we then take the moving average of parameters of the feature extractor as the weight generator. That is,
\begin{equation} \label{eq:ema}
   \phi'_t = \alpha \phi'_{t-1} + (1 - \alpha) \phi_{t}   
\end{equation}
 where $\phi'_t$ is the parameter of the weight generator at time $t$ and $\phi_{t}$ is the parameters of the feature extractor updated by SGD, $\alpha$ is the momentum hyperparameter. 

\begin{figure}[t]
   \centering
   \includegraphics[width=1.0\linewidth]{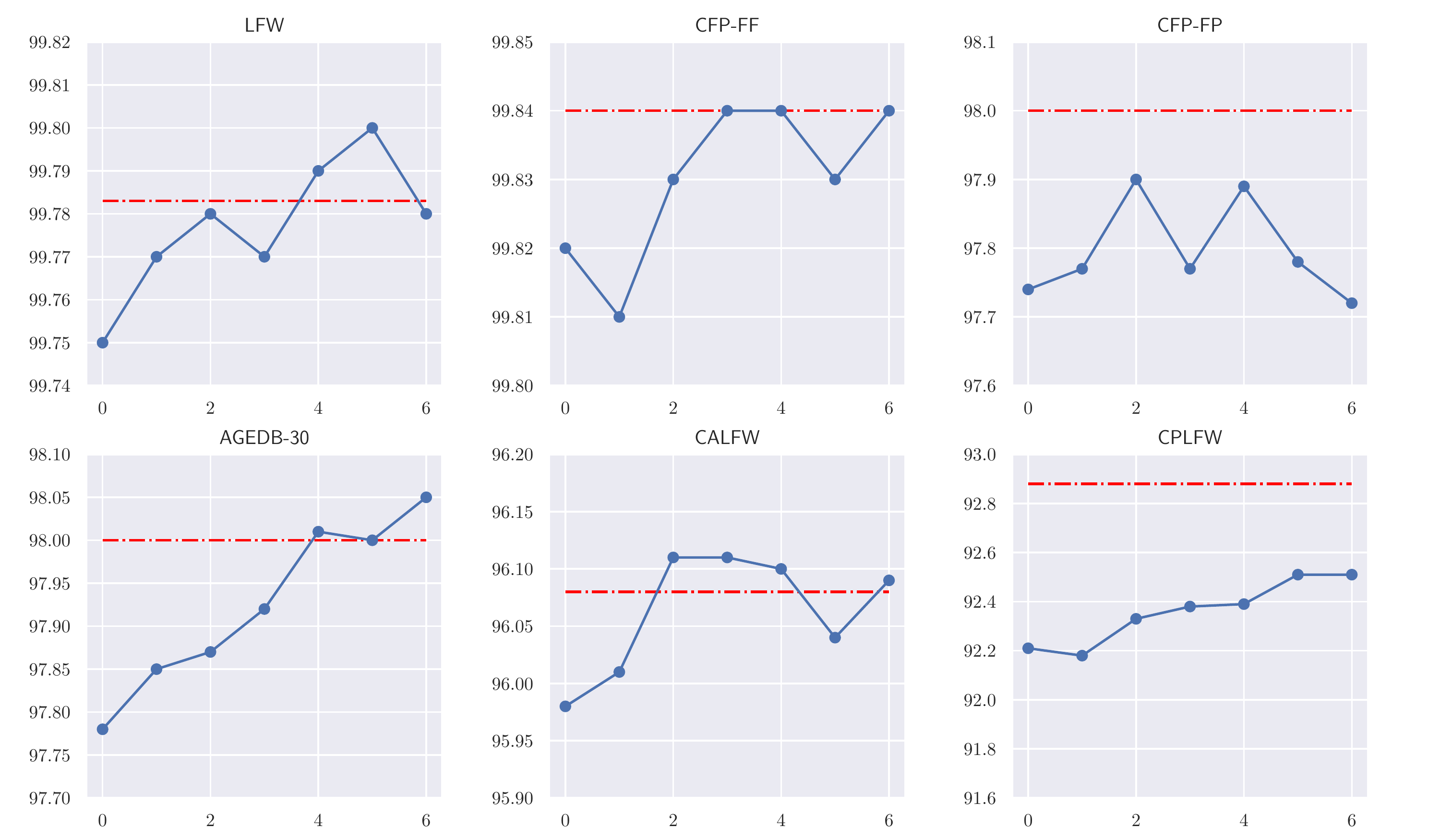}
   \caption{\textbf{The effects of the queue size}. The $x$-axis is the queue size $2048\times 2^x$. The $y$-axis is the verification accuracies on six benchmarks. The \textcolor{red}{red} dotted line is the CosFace baseline using all classes. Models are trained on MS1MV2 with ResNet50 backbone. Experiments are repeated 3 times and we report the mean. \label{fig:queue_size}}
\end{figure}

\section{Experiment}
\subsection{Experimental Setup}
\textbf{Training Dataset.} To evaluate different aspects of the proposed method, two large-scale datasets are used respectively. For training under clean dataset with balanced classes, we use \textbf{MS1MV2}\cite{deng2019arcface}. It is a cleaned version of MS-1M-Celeb\cite{guo2016ms}. This dataset contains images of celebrities which consists of 85K identities and 5.8M images. It has balanced classes, each identity has about 100 images on average. For training with the long-tailed dataset, we use the training set of \textbf{Megaface Challenge 2 (MF2)}\cite{nech2017level}. This is one of the largest datasets publicly available. It has 672K identities and 4.7M images, each identity has 7 images on average. Unlike MS1MV2, the images of MF2 are collected from Flickr and most of them are common people. In MF2, the classes are long-tail distributed, 88.42\% of identities have less than 10 images. 

\textbf{Training Settings.} Two ResNet\cite{he2016deep} backbones with different depth are used for feature extraction: ResNet50 and ResNet101. For fair comparison, we adopt the backbone improvements (remove the first pooling operation, change ReLU\cite{nair2010rectified} to PReLU\cite{he2015delving}, etc.) proposed in \cite{deng2019arcface}. The models are trained using SGD with momentum 0.9 and weight decay 0.0001. The batch size is 512. For the CosFace baseline, the initial learning rate is 0.1. For DCQ, it is 0.06. Both methods follow the same learning rate schedule. For MS1MV2, the learning rate is decreased 10x at epoch 8, 16, and 18. The total epoch is 20. For MF2, the learning rate is decreased at epoch 25, 35, 38, for a total of 40 epochs.

\textbf{Hyperparameters.} The feature dimension of all models is 512. For our baseline model, we use the same hyperparameters as the ones proposed in the CosFace\cite{wang2018cosface}. The scale and the margin in the loss function are 64 and 0.35. For our DCQ model, we find that smaller scale and margin result in better performance (not the case for our baseline model), therefore we use scale 50 and margin 0.3. The momentum $\alpha$ in Equation~\ref{eq:ema} is set to 0.999.

\textbf{Data Preprocessing.} We use MTCNN\cite{zhang2016joint} to locate 5 facial landmarks and use them to normalize the face to a canonical position. The normalized face is then cropped and resized to 112x112. During training, the RGB values are normalized to the range (-1, 1) by subtracting 127.5 and then dividing by 128. For training on MS1MV2, we use only flipping augmentation with probability 50\%. For MF2, besides flipping, we also use monochrome augmentation with probability 20\%.

\textbf{Testing Settings.} During testing, following \cite{deng2019arcface} features of the original image and the flipped image are averaged as the final testing feature. Cosine distance of the features is used. A single crop is used for all tests. We tests our models on standard verification benchmarks including LFW\cite{LFWTech}, CFP-FF, CFP-FP\cite{sengupta2016frontal}, AGEDB-30\cite{moschoglou2017agedb}, CALFW\cite{DBLP:journals/corr/abs-1708-08197} and CPLFW\cite{zheng2018cross}. Moreover, the model is evaluated on the test set of the Megaface Challenge 2. The test set includes two tasks: identification and verification. For the identification task, it has 1M distractors (disjoint with the MF2 training set) in the gallery set and 100K images from FaceScrub\cite{ng2014data} as the probe set.

\subsection{Exploratory Experiments}
\textbf{Dynamic Class Queue works as well as using all classes.} One of the most important questions in this work is how does DCQ perform compared to baselines using all classes during training? To answer this question, we compare DCQ with baselines by training on the MS1MV2 dataset. Although MS1MV2 has identities as large as 85K, it is still possible to fit the whole FC layer into GPU memory and therefore suitable for training our baselines without compromise. Moreover, this dataset is class-balanced and allows us to focus on the approximation ability of the dynamic class queue without considering the long-tail effect (which will be discussed in MF2 based experiments).

For fast experimentation, we use ResNet50 as the backbone. The performance of DCQ with different queue sizes is shown in Figure~\ref{fig:queue_size}. As can be seen, the recognition performance steadily improves when the queue size increases from 2048 to 16384. Notice that when the queue size is 8192, it has about 10\% of all classes (85,742) and already performs as well as the CosFace baseline using all classes. By further increasing the queue size, DCQ is able to outperform the baseline method in LFW, AGEDB-30, and CALFW while performs slightly worse (within 0.3\%) in CFP-FP and CPLFW. This is likely due to the different requirements of intra- and inter- variance in different benchmarks. We leave more in-depth investigations for future works. Nevertheless, DCQ achieves a good balance between accuracy and efficiency. As we will show, the benefits of weight generation outweigh the drawbacks of partial classes and DCQ achieves superior results in the long-tailed dataset.

\textbf{Dynamic Class Queue is hardware friendly.}
One major benefit of the proposed method is in terms of computational cost and memory consumption. We show the GPU memory used in one card and seconds per batch in Table~\ref{tab:memory_compute}. Both methods are trained on the MF2 dataset. During training, the CosFace has 642,962 classes and the DCQ has queue size 65,536. As we will show, using 10\% of classes can outperform the baseline in the MF2 dataset. ResNet50 is used as the backbone and batch size is 512. Data loading time is removed. Code runs on a single server with 8 V100 32G GPU cards. 

\begin{table}[h]
   \centering
   \footnotesize
   \caption{\textbf{Time and memory comparison}. \label{tab:memory_compute}}
   \begin{tabular}{| l | c c c|}
      \hline
      model & Classes & Time (Sec/batch) & Memory (MB) \\
      \hline\hline
      CosFace & 642,962 & 0.291 & 18,621\\
      DCQ & 65,536 & 0.245 & 8,435\\
      \hline
   \end{tabular}
\end{table}

\textbf{DCQ works well for the long-tailed dataset.} We have just verified the effectiveness of DCQ on a clean and balanced dataset, MS1MV2. Now, we will let DCQ go through the real test: MF2. MF2 is by far the public dataset with the largest number of identities: 672K, which has about 8x more identities than MS1MV2 (85K). Moreover, it is a long-tailed dataset, over 80\% identities have less than 10 images. It is representative of datasets used in real applications. Models are trained with ResNet50 backbone and then tested on the test set of the Megaface Challenge2.

\begin{table}[h]
   \centering
   \footnotesize
   \caption{\textbf{Comparison on training with long-tailed dataset: MF2} \label{tab:long-tail}}
   \resizebox{\columnwidth}{!}{
   \begin{tabular}{|l | c c| c c c|}
      \hline
      model & MF2 Id. (\%) & MF2 Ver.(\%) & LFW & AGEDB-30 & CPLFW\\
      \hline\hline
      CosFace-h9 & 77.10 & 86.75 & 99.58 & 90.35 & 88.60 \\
      CosFace-all & 77.24 & 87.56 & 99.60 & 90.27 & 88.52\\
      DCQ & 77.99 & 89.28 & 99.58 & 91.07 & 90.12 \\
      \hline
   \end{tabular}
   }
\end{table}

We compare with two baselines, CosFace-h9 is trained only on the head classes with instances $\ge 9$, which has 100K classes and 2.2M images. CosFace-all is trained with all classes. Our DCQ model uses 10\% of all classes. The results are shown in Table~\ref{tab:long-tail}. Notice that simply removing all tail classes degrades the final performance. On the other hand, by adopting the proposed DCQ method, with only 10\% of all classes, it is able to outperform the baseline 0.75\% in identification and 1.72\% in verification. This demonstrates the effectiveness of DCQ in the long-tailed dataset. Moreover, the improvement is consistent in LFW, AGEDB-30, and CPLFW benchmarks.

\textbf{Momentum update is very important.} The central design of this work is to maintain a queue that contains class weights generated on-the-fly at different iterations during training. To investigate the importance of smoothing the feature drift, we use different momentums $\alpha$ to train our DCQ model with the ResNet50 backbone. Intuitively, when $\alpha$ is as small as $0$, there is no feature smoothing between different iterations. On the other hand, when $\alpha$ is as large as $1$, the feature extractor will remain unchanged and cant reflect the current feature distribution. Since the feature drift problem is more evident when the queue is large, we use queue size 65,536 and train the model on MS1MV2. 

\begin{figure}[t]
   \centering
   \footnotesize
   \includegraphics[width=0.8\linewidth]{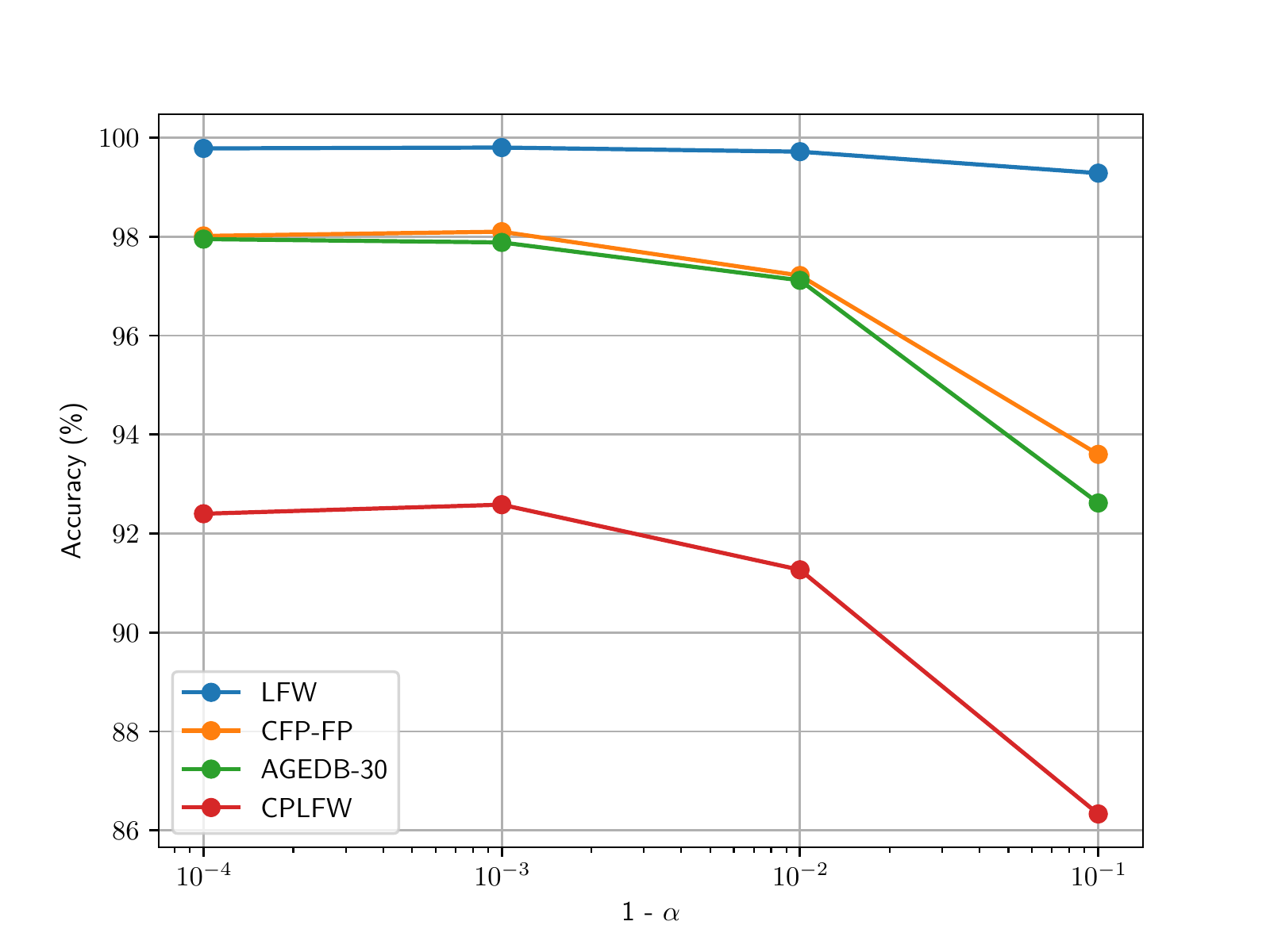}
   \caption{\textbf{The effects of momentum update with different momentum $\alpha$}. The $x$-axis is 1 - $\alpha$ and in log scale. The $y$-axis is the verification accuracies. The DCQ model is trained on MS1MV2 with ResNet50 backbone. The queue size 65,536. \label{fig:moving_average}}
\end{figure}

The performance on LFW, CFP-FP, AGEDB-30, and CPLFW with different momentums are presented in Figure~\ref{fig:moving_average}. As can be seen, as the moving average $\alpha$ increases from $0.9$ (the right side of the figure), the accuracies are consistently improved on all datasets and saturate at $0.999$ (the left side of the figure). This demonstrates the importance of momentum update for mitigating the feature drift. Note that the momentum has different impacts on different datasets. For LFW, the influence is marginal while for other datasets the improvement can be over 6\%. This implies that LFW is a relatively easy testing benchmark compared to other datasets and does not require very strong features.

\textbf{Instance-based is better than class-based sampling for DCQ.} For long-tailed datasets, one common practice is to adopt class-based sampling, that is, all classes are sampled with equal probability. This is in contrast to instance-based sampling where all images are sampled equally and therefore tail classes with more images are more likely to be sampled. Interestingly, the proposed DCQ model performs better with instance-based sampling even in long-tailed dataset as shown in Table~\ref{tab:sampling}. One possible explanation is that class-balanced sampling is good for learning the classifier (e.g. the FC layer) since all classes are equally sampled without bias. However, it is bad for representation learning which requires diversity in the training data, and increasing the probability of sampling tail classes just means the same data are repeated several times. Since the DCQ model does not need to learn the classifier which is directly generated, it is beneficial to use instance-based sampling.

\begin{table}[h]
   \centering
   \footnotesize
   \caption{\textbf{Comparisons between instance- and class- sampling}, \label{tab:sampling}}
      \resizebox{\columnwidth}{!}{
   \begin{tabular}{|l | c  c | c c c |}
      \hline
      model & MF2 Id.(\%) & MF2 Ver.(\%) & LFW & AGEDB-30 & CPLFW \\
      \hline\hline
      DCQ-cls & 76.78 & 88.44 & 99.51 & 90.55 & 89.42 \\
      DCQ-ins & 77.99 & 89.28 & 99.58 & 91.07 & 90.12  \\
      \hline
   \end{tabular}
   }
\end{table}

\textbf{DCQ converges as fast as CosFace in the balanced dataset and faster in the long-tailed dataset.} Since the class queue is randomly assembled and changes after each iteration, one concern is about the convergence speed of DCQ. To investigate this problem, we train ResNet50 backbones with CosFace and DCQ using the same training settings (optimizer, number of epochs, and learning rate schedule). These settings are adopted following the best practice of training CosFace, we do no tune them for DCQ. For DCQ, we use 10\% classes.
The performance of each epoch is shown in Figure. As can be seen, DCQ converges as fast as CosFace in MS1MV2 (balanced dataset) and faster in MF2 (long-tailed dataset). This is beneficial for the long-tailed dataset in the wild.

\begin{figure}[h]
   \centering
   \includegraphics[width=0.9\linewidth]{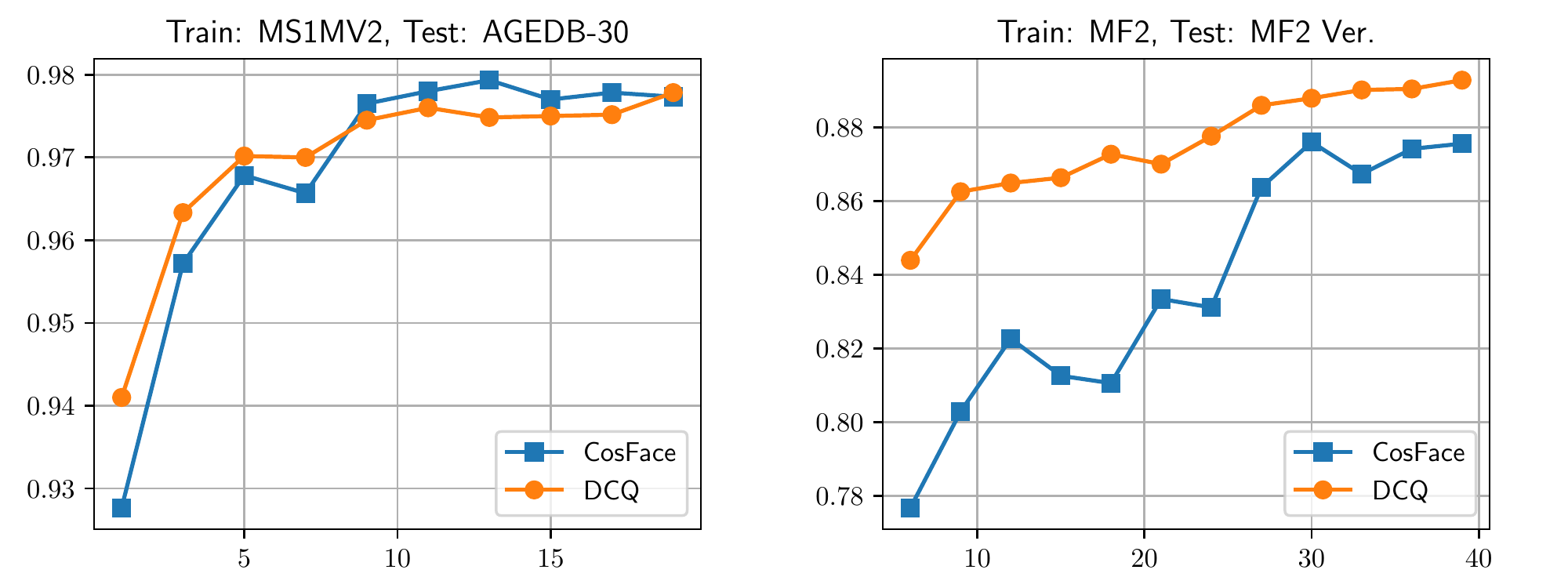}
   \caption{\textbf{Convergence comparisons}. The $x$-axis is the epoch. The $y$-axis is the verification accuracy. The left is trained on MS1MV2 and tested on AGEDB-30 while the right is trained on MF2 and tested on the verfication benchmark of MF2.\label{fig:convergence}}
\end{figure}

\section{State-of-the-art Comparisons}

In this section, we compare DCQ with state-of-the-art methods to give readers an idea of how good the proposed method is in the face recognition field. Yet we emphasize that achieving the best recognition performance is not the focus of this work. Instead, we aim for a new paradigm for face recognition in the wild with large-scale datasets. The proposed method is general enough to easily combine with other methods (ArcFace loss, Curriculum Learning, etc.) to further boost the performance. Without bells and whistles, the proposed method is already competitive with other state-of-the-art methods.

\textbf{Comparisons on MS1MV2.} We train models on the MS1MV2 dataset and test them on standard benchmarks. Following most SOTA methods, we use ResNet101 as the backbone. The results are shown in Table~\ref{tab:sota-ms1mv2}. As can be seen, our method performs on par with most SOTA methods. In particular, Partial FC is a recently proposed method that reduces computations by randomly sampling classes at each iteration and bears similarities with our method. The major difference is that in PartialFC, class weights are learned via SGD while we directly generate them via one-shot learning. We achieve a similar performance as the partial FC. This implies that directly generates class weights on-the-fly works as well as learning them via SGD. The extra benefit of DCQ is that it requires less GPU memory during training since Partial FC needs to store all class weights in the GPU through model parallel.

\begin{table}[h]\centering
   \footnotesize
   \caption{\textbf{SOTA comparisons on models trained with MS1MV2.}\label{tab:sota-ms1mv2}}
   \resizebox{\columnwidth}{!}{
   \begin{tabular}{|l | c c c c c|}
      \hline
      model & LFW & CFP-FP & AGEDB-30 & CALFW & CPLFW \\
      \hline\hline
      CosFace\cite{wang2018cosface} & 99.43 & - & - & 90.57 & 84.00 \\
      ArcFace\cite{deng2019arcface} & 99.82 & 98.27 & - & 95.45 & 92.08 \\
      GroupFace\cite{kim2020groupface} & 99.85 & 98.63 & 98.28 & 96.20 & 93.17 \\
      CurricularFace\cite{huang2020curricularface} & 99.80 & 98.37 & 98.32 & 96.20 & 93.13 \\
      PartialFC-r1.0\cite{an2020partial} & 99.83 & 98.51 & 98.03 & 96.20 & 93.10 \\
      PartialFC-r0.1\cite{an2020partial} & 99.82 & 98.60 & 98.13 & 96.12 & 92.90 \\
      \hline\hline
      CosFace(ours) & 99.78 & 98.38 & 98.22 & 96.20 & 93.15 \\
      \textbf{DCQ(ours)} & \textbf{99.80} & \textbf{98.44} & \textbf{98.23} & \textbf{96.07} & \textbf{92.87} \\
      \hline
   \end{tabular}
   }
\end{table}

\textbf{Comparisons on MF2.} We train models on the MF2 dataset and test them on the test set of MF2. For a fair comparison with other methods, ResNet50 is used as the backbone. The results are shown in Table~\ref{tab:sota-mf2}. As can be seen, the proposed DCQ outperforms most SOTA methods. NRA+CD achieves better performance than our method with multi-stage training and careful noise reduction in the MF2 dataset. In fact, the NRA+CD is orthogonal to our method and should further improve the performance when combined together.

\begin{table}[h]
   \centering
   \footnotesize
   \caption{\textbf{SOTA comparisons on models trained with MF2.}\label{tab:sota-mf2}}
   \begin{tabular}{|l | c c|}
         \hline
         model & MF2 Id.(\%) &  MF2 Ver.(\%) \\
         \hline\hline
         3DiVi & 57.05 & 66.46 \\
         NEC & 62.12 & 66.85 \\
         RangeLoss & 69.54 & 82.67 \\
         SphereFace & 71.17 & 84.22 \\
         GRCCV & 75.77 & 74.84 \\
         Yang Sun & 75.79 & 84.03 \\
         CosFace\cite{wang2018cosface} & 74.11 & 86.77 \\
         NRA+CD\cite{zhong_unequal-training_2019} & 80.02 & 89.93 \\
         \hline\hline
         CosFace(Ours) & 77.24 & 87.56 \\
         \textbf{DCQ(Ours)} & \textbf{77.99} & \textbf{89.28} \\
         \hline
   \end{tabular}
\end{table}

\section{Conclusion}
In this work, we propose a new framework for training with large-scale datasets in the wild. Specifically, two challenging problems, i.e., hardware constraints and long-tailed classes, are solved simultaneously in a unified way via the dynamic class queue. In essence, the proposed method dynamically selects classes for classification which reduces the computing cost and dynamically generates class weights which saves memory and overcomes the long-tail issue. The DCQ model is empirically validated on two large-scale datasets where it achieves similar performance as using all classes in the balanced dataset and better performance in the long-tailed dataset.

{\small
\bibliographystyle{ieee_fullname}
\bibliography{egbib}
}

\end{document}